\documentclass[conference]{IEEEtran}
\IEEEoverridecommandlockouts
\usepackage{cite}
\usepackage{url}
\usepackage{amsmath,amssymb,amsfonts}
\usepackage{algorithmic}
\usepackage{graphicx}
\usepackage{textcomp}
\usepackage{xcolor}
\usepackage{float}
\usepackage[utf8]{inputenc}
\usepackage[T5]{fontenc}
\usepackage{babel}
\usepackage{multirow}
\usepackage{makecell}
\usepackage{fancyvrb}
\usepackage{xcolor}
\usepackage{mdframed}
\usepackage{hyperref}
\usepackage{tabularx}

\definecolor{shadecolor}{rgb}{1,.8,.3}
\babelprovide[import]{vietnamese}
\def\BibTeX{{\rm B\kern-.05em{\sc i\kern-.025em b}\kern-.08em
    T\kern-.1667em\lower.7ex\hbox{E}\kern-.125emX}}
\makeatletter
\newcommand{\linebreakand}{%
  \end{@IEEEauthorhalign}
  \hfill\mbox{}\par
  \mbox{}\hfill\begin{@IEEEauthorhalign}
}
\makeatother

\begin{document}

\title{Investigating Recent Large Language Models for Vietnamese Machine Reading Comprehension\\
}

\author{
\IEEEauthorblockN{Anh Duc Nguyen, Hieu Minh Phi, Anh Viet Ngo, Long Hai Trieu, Thai Phuong Nguyen}
\IEEEauthorblockA{\textit{University of Engineering and Technology - Vietnam National University}\\
(22022504, 21020200, 22022508, thlong, thainp)@vnu.edu.vn}
}

\maketitle

\begin{abstract}
Large Language Models (LLMs) have shown remarkable proficiency in Machine Reading Comprehension (MRC) tasks; however, their effectiveness for low-resource languages like Vietnamese remains largely unexplored. In this paper, we fine-tune and evaluate two state-of-the-art LLMs: Llama 3 (8B parameters) and Gemma (7B parameters), on ViMMRC, a Vietnamese MRC dataset. By utilizing Quantized Low-Rank Adaptation (QLoRA), we efficiently fine-tune these models and compare their performance against powerful LLM-based baselines. Although our fine-tuned models are smaller than GPT-3 and GPT-3.5, they outperform both traditional BERT-based approaches and these larger models. This demonstrates the effectiveness of our fine-tuning process, showcasing how modern LLMs can surpass the capabilities of older models like BERT while still being suitable for deployment in resource-constrained environments. Through intensive analyses, we explore various aspects of model performance, providing valuable insights into adapting LLMs for low-resource languages like Vietnamese. Our study contributes to the advancement of natural language processing in low-resource languages, and we make our fine-tuned models publicly available at: \url{https://huggingface.co/iaiuet}.
\end{abstract}

\begin{IEEEkeywords}
Large language models, machine reading, comprehension, fine-tuning models
\end{IEEEkeywords}

\begin{figure}[ht!]
\centering

\begin{mdframed}[
  linecolor=black,
  linewidth=1pt,
  topline=true,
  bottomline=true,
  middlelinecolor=black, 
  leftline=true,
  rightline=true,
  backgroundcolor=white,
  skipabove=\topsep,
  skipbelow=\topsep,
]
\noindent
\textbf{Reference:} 
\\Bố đi câu về, không một lần nào là chúng tôi không có quà.

Mở thúng câu ra là cả một thế giới dưới nước: cà cuống, niềng niễng đực, niềng niễng cái bò nhộn nhạo. Hoa sen đỏ, nhị sen vàng tỏa hương thơm lừng. Những con cá sộp, cá chuối quẫy tóe nước, mắt thao láo...




(\textit{When Dad went fishing, he never came back without gifts for us.\\
Opening the fishing basket was like revealing a whole underwater world: giant water bugs, male and female damselflies wriggling around. Red lotus flowers, yellow lotus stamens spreading a fragrant aroma. Big fish and snakehead fish splashing water, their eyes bulging...
})
\end{mdframed}

\begin{mdframed}[
  linecolor=black,
  linewidth=1pt,
  topline=true,
  bottomline=true,
  middlelinecolor=black, 
  leftline=true,
  rightline=true,
  backgroundcolor=white,
  skipabove=\topsep,
  skipbelow=\topsep
]
\noindent
\textbf{Question:} Quà của bố có những gì sau những buổi đi câu về? (\textit{What are dad's gifts after returning from fishing trips?})

\end{mdframed}
\begin{mdframed}[
  linecolor=black,
  linewidth=1pt,
  topline=true,
  bottomline=true,
  middlelinecolor=black, 
  leftline=true,
  rightline=true,
  backgroundcolor=white,
  skipabove=\topsep,
  skipbelow=\topsep
]
\noindent
\textbf{Options:} \\
A. Cà cuống, niềng niễng, hoa sen, cá sộp và cá chuối. (\textit{Water bugs, damselflies, lotus, big fish and snakehead fish}) \\
B. Con chim, con châu chấu và con bướm. (\textit{Bird, grasshopper and butterfly})\\
C. Con khỉ, con thỏ, con cò hương và con khướu. (\textit{Monkey, rabbit, stork and babbler})\\
D. Con xập xành, con muỗm và con dế. (\textit{Giant crickets, mole crickets and male crickets})    
\end{mdframed}
\caption{An example of MRC task (English translation is in italics).}
\label{fig:mrc_example}
\end{figure}

\section{Introduction}
Machine Reading Comprehension (MRC) has emerged as a critical area of research in Natural Language Processing (NLP), pushing the boundaries of machine intelligence in understanding and processing human language. MRC tasks, which involve the complex interplay between a passage, a question, and an answer, challenge machines to comprehend text and accurately respond to queries based on that understanding \cite{dzendzik-etal-2021-english,9664302}. This capability is fundamental to advanced NLP applications and serves as a benchmark for language understanding and reasoning abilities in artificial intelligence systems. For instance, consider the example shown in Fig. \ref{fig:mrc_example}, which illustrates a typical MRC task. The MRC systems must understand the passage and accurately identify the correct answer from the given options. Such tasks not only test the system's comprehension abilities but also its capacity for reasoning and contextual understanding \cite{9664302}.

In recent years, large language models (LLMs) have demonstrated remarkable proficiency across various NLP tasks, including MRC, achieving state-of-the-art (SOTA) results \cite{Zhao2023ASO, Zhou2023ACS, Bommasani2021FoundationModels}.

However, the performance of these models often diminishes when scaled down to smaller parameter sizes, particularly for languages with fewer resources than English \cite{Gholami2023DoGL}. This challenge is particularly evident in Vietnamese, a language with relatively fewer NLP resources compared to English. The introduction of the ViMMRC dataset \cite{vimmrc} has provided a standardized benchmark for evaluating and improving Vietnamese MRC systems. However, the effectiveness of SOTA LLMs on this dataset remains unexplored, highlighting the need for improvement in Vietnamese MRC capabilities.

To address this gap, our study focuses on evaluating and fine-tuning two cutting-edge LLMs, i.e., Llama 3 (8B parameters) \cite{llama3modelcard} and Gemma (7B parameters) \cite{team2024gemma}, on the ViMMRC dataset. We aim to demonstrate the necessity and effectiveness of fine-tuning LLMs to enhance their performance in Vietnamese MRC tasks. Our results show that fine-tuned base LLMs achieve significantly higher accuracy compared to their pre-fine-tuned versions. Notably, our fine-tuned models, despite having relatively small parameter counts, outperform much larger models such as GPT-3 and GPT-3.5, demonstrating the efficiency and effectiveness of our fine-tuning process. This research not only contributes to the advancement of Vietnamese NLP but also provides valuable insights into the adaptability of LLMs for specific languages and tasks. Our findings have potential implications for educational technology, information retrieval systems, and other AI-driven applications in Vietnamese-language contexts.

The main contributions of this paper are as follows:
\begin{itemize}
\item Several SOTA LLM models are evaluated and finetuned using the Quantized Low-Rank
Adaptation (QLoRA) on a Vietnamese MRC task, which are efficient in terms of computational resources due to fewer parameters but achieve SOTA performance and outperform several powerful baselines.
\item Intensive analyses are conducted on the Vietnamese MRC task, which we reveal opportunities and challenges for adapting LLMs to low-resource languages.
\item Our finetuned models are publicly released for community to further advance LLM-based MRC task as well as for Vietnamese NLP applications.
\end{itemize}

\begin{table}[t!]
\caption{Statistics of dataset ViMMRC}
\begin{center}
\resizebox{0.48\textwidth}{!}
{
\begin{tabular}{l|rrrrrr}

\hline
\textbf{Grade} & \textbf{1} & \textbf{2} & \textbf{3} & \textbf{4} & \textbf{5} & \textbf{All} \\
\hline
\textbf{Number of texts} & 10 & 70 & 188 & 99 & 120 & 417 \\
\textbf{Vocabulary size} & 595 & 3,325 & 4,666 & 5,006 & 5,702 & 10,099 \\
\textbf{Number of questions} & 60 & 514 & 759 & 709 & 741 & 2,783 \\
\hline
\end{tabular}
}
\end{center}
\label{tab:vimmrc_stat}
\end{table}

\section{Models}

In order to evaluate recent LLMs on Vietnamese MRC task, we selected several models based on their advanced capabilities and public accessibility including: Llama 3 \cite{llama3modelcard} and Gemma \cite{team2024gemma}. 
First, these models are openly available.
Second, both Llama 3 and Gemma offer base and instruction-tuned variants, as well as versions with reduced parameter counts. The availability of smaller model sizes is particularly advantageous for our experimental setup, as it enables efficient fine-tuning on the ViMMRC dataset using limited computational resources.

\begin{table}[t]
\centering
\caption{Chat template to prompt the model for answering multiple-choice questions}
\label{tab:chat}
\begin{tabular}{|p{0.9\linewidth}|}
\hline
\textbf{user}: Use the following reference to answer the given multiple-choice question:\\
Reference: \textit{\textless Reference\textgreater} \\
Question: \textit{\textless Question\textgreater} \\
A. \textit{\textless Option 1\textgreater} \\
B. \textit{\textless Option 2\textgreater} \\
C. \textit{\textless Option 3\textgreater} \\
D. \textit{\textless Option 4\textgreater} \\
\hline
\textbf{model}: 
\textit{\textless Answer\textgreater} (A, B, C, or D) \\
\hline
\end{tabular}
\end{table}

\begin{table}[t]
\caption{Hyperparameters}
\begin{center}
{
\begin{tabular}{|l|r|}
\hline
\textbf{Hyper-parameter} & \textbf{Value} \\
\hline
Batch Size & 64  \\
Epochs & 3  \\
Sequence Length & 8192  \\
Warm-up Steps & 5  \\
Weight Decay & 0.01 \\
Optimizer & AdamW \\
Learning rate scheduler & cosine\\
\hline
\end{tabular}}
\label{tab:hyperparameters}
\end{center}
\end{table}

\section{Experimental Settings}
\label{sec3}

\subsection{Llama 3}
Llama 3 \cite{llama3modelcard} 
is published by MetaAI and available in 8 billion (8B) and 70 billion (70B) parameter versions. Building on the foundation laid by earlier Llama models~\cite{touvron2023llama2openfoundation}, Llama 3 is designed to achieve high-quality performance through extensive pre-training data rather than sheer parameter count. Compared to its predecessor Llama 2, Llama 3 features significant upgrades, including a tokenizer capable of handling 128k tokens and training on a massive 15 trillion tokens, which is seven times the data used for Llama 2. This dataset includes 5\% non-English data, ensuring a diverse and comprehensive training process. Llama 3 also undergoes high-quality filtering to enhance its overall performance, making it a powerful tool for various NLP applications. We employ the 8B version of Llama 3 in our study, and will elaborate on its performance in Section \ref{sec3}.

\subsection{Gemma}
Gemma \cite{team2024gemma}
is a family of lightweight, SOTA open models from Google DeepMind, derived from the Gemini research and technology. Available in 2 billion (2B) and 7 billion (7B) parameter versions, Gemma demonstrates robust performance across various academic benchmarks for language understanding, reasoning, and safety. Notably, the 7B model is optimized for efficient deployment on GPU and TPU, while the 2B model is designed for CPU and on-device applications. Gemma models are trained on up to 6 trillion tokens, primarily from English web documents, mathematics, and code. This extensive pre-training, combined with a high-quality filtering process, ensures the models' strong generalist capabilities and SOTA performance in numerous text domains. We utilize the 7B version of Gemma in our study due to its feasibility on GPU, and will elaborate on its performance in Section \ref{sec3}.

\subsection{Dataset}

In this paper, we utilize the ViMMRC dataset
\cite{vimmrc}, a pioneering corpus for Vietnamese multiple-choice machine reading comprehension (MRC). It comprises 2,783 question-answer pairs extracted from 417 texts intended for students from 1st to 5th grade. The distribution of texts for different grades is demonstrated in Table \ref{tab:vimmrc_stat}. 
Each text in the ViMMRC is curated to test various levels of comprehension, from basic factual recall to more complex reasoning, reflecting the diverse challenges in natural language understanding. By focusing on multiple-choice questions, the dataset allows for standardized evaluation of MRC systems, facilitating benchmarking and comparison across different models and approaches. 
The ViMMRC dataset was partitioned into: 
1,975 pairs for training set, 294 pairs for development set, and 514 pairs for test set.

\subsection{Fine-tune settings}
In this study, we employ the Quantized Low-Rank Adaptation (QLoRA) method \cite{qlora} to fine-tune our language models. QLoRA is a parameter-efficient method for adapting pre-trained language models by utilizing quantization and low-rank adaptation. This technique reduces the computational cost and memory footprint, making it feasible to handle large models on low computational resources.

We adopt the Unsloth template \cite{Unsloth} for our QLora implementation. Following their recommendations, we set the learning rates to $5e-5$ for Gemma and $2e-4$ for Llama, with additional hyperparameters detailed in Table \ref{tab:hyperparameters}. 

Our implementation employs a structured chat template (Table \ref{tab:chat}) to prompt the model for answering multiple-choice questions based on a given reference. The template follows a consistent format to ensure clarity and effectiveness. 
The chat template consists of a clear distinction between \textit{user input} and \textit{model output}
The user input section provides the necessary context for the model, including the reference text and the multiple-choice question with options labeled A, B, C, and D. The model's response is expected in the form of selecting one of these options. This structured format helps streamline the interaction and ensures that the model's responses are aligned with the user's expectations.

\subsection{Baseline models}

To evaluate the effectiveness of our fine-tuned models, we compare the performance of our models against several powerful baseline models as follows.

\begin{table}[t]
\caption{Results of our fine-tuned models on ViMMRC}
\begin{center}
{
    \begin{tabular}{l|r|c}
    \hline
    \textbf{Model} & \textbf{Size} & \textbf{Accuracy} \\
    \hline
    BERTology  & 179M & 82.29 \\
    Llama 2    & 7B & 33.46 \\
    BLOOMZ     & 7.1B & 79.96 \\
    GPT-3      & 175B & 77.04 \\
    GPT-3.5    & 175B & 82.88 \\
    \hline
    Llama 3$_{\text{Instructed}}$  & 8B & 84.21 \\
    Gemma$_{\text{Instructed}}$  & 7B & 84.94 \\
    Llama 3$_{\text{Base}}$  & 8B & 84.93 \\
    Gemma$_{\text{Base}}$ & 7B & \textbf{87.11} \\
    \hline
    \end{tabular}
}
\label{tab:baseline}
\end{center}
\end{table}

\begin{table}[t]
\caption{Results on ViMMRC with finetuned models (Without Finetuning: the original models; Finetuned: our models after finetuned; Better scores are in bold.)}
\begin{center}
    \begin{tabular}{l|c|c}
    \hline
    \textbf{Model} & \textbf{Without Finetuning} & \textbf{Finetuned} \\
    \hline
    Llama 3$_{\text{Instructed}}$ & 56.80 & \textbf{84.21} \\
    Gemma$_{\text{Instructed}}$ & 80.39 & \textbf{84.94} \\
    \hline
    Llama 3$_{\text{Base}}$ & N/A & \textbf{84.93} \\
    Gemma$_{\text{Base}}$ & N/A & \textbf{87.11} \\
    \hline
    \end{tabular}
\label{tab:finetune}
\end{center}
\end{table}

\begin{table*}[t!]
\centering
\caption{\label{tab:bias} Sample of bias mitigation}

\begin{tabularx}{\textwidth}{l|X|X}
\hline
& \textbf{Sample} & \textbf{English translation} \\
\hline
\textbf{Passage} &
Sau một chuyến đi xa, người ông mang về bốn quả đào. Ông bảo vợ và
các cháu: Quả to này xin phần bà. Ba quả nhỏ hơn phần các cháu. Bữa cơm chiều hôm ấy, ông hỏi các cháu: Thế nào, các cháu thấy đào có ngon không? Cậu bé Xuân nói: Đào có vị rất ngon và mùi thật là thơm. Cháu đã đem hạt trồng vào một cái vò. Chẳng bao lâu, nó sẽ mọc thành một cây đào to đấy, ông nhỉ? Mai sau cháu sẽ làm vườn giỏi. - Ông hài lòng nhận xét. Cô bé Vân nói với vẻ tiếc rẻ: Đào ngon quá, cháu ăn hết mà vẫn thèm. Còn hạt thì cháu vứt đi rồi. Ôi cháu của ông còn thơ dại quái. Thấy Việt chỉ chăm chú nhìn vào tấm khăn trải bàn, ông ngạc nhiên hỏi: Còn Việt, sao cháu chẳng nói gì thế? Cháu ấy ạ? Cháu mang đào cho Sơn. Bạn ấy bị ốm. Nhưng bạn ấy không
muốn nhận. Cháu đặt quả đào lên trên giường rồi trốn về. Cháu là người có tấm lòng nhân hậu! - Ông lão thốt lên và xoa đầu đứa cháu nhỏ. &

After a long trip, the grandfather brought back four peaches. He told his wife and grandchildren: This big peach is for your grandmother. The three smaller ones are for you children. At dinner that evening, the grandfather asked his grandchildren: How did you like the peaches? Xuan said: The peaches were very delicious and smelled wonderful. I planted the pit in a pot. Soon, it will grow into a big peach tree, right, Grandpa? You will become a great gardener when you grow up, - the grandfather remarked, pleased. Little Van, with a regretful expression, said: The peach was so delicious that I ate it all and still wanted more. I threw the pit away. Oh, my little one is still so naive. Seeing Viet staring intently at the tablecloth, the grandfather asked in surprise: And Viet, why haven't you said anything? Me? I gave my peach to Son. He's sick. But he didn't want to take it. So, I placed the peach on his bed and sneaked away. You have a kind heart! - the old man exclaimed, patting his little grandson on the head.\\
\hline
\textbf{Question} & Trong ba đứa cháu, ai là đứa cháu được ông nhận xét là còn thơ dại? 
&
In three children, who is claimed as innocent? 
\\
\hline
\textbf{Answers} & A. Vân \quad B. Việt \quad C. Xuân \quad D. Cả 3 bạn trên
&
A. Van \quad B. Viet \quad C. Xuan \quad D. All of the above
\\
\hline
\textbf{Predictions} & \multicolumn{2}{l}{
\textbf{Correct answer}: A \quad 
\textbf{BERTology}: D \quad
\textbf{Llama 3}: A \quad
\textbf{Gemma}: A
} \\
\hline
\multicolumn{3}{c}{}\\
\multicolumn{3}{p{0.95\linewidth}}{\textit{Notes:} Compared to the BERT-based model, our fine-tuned models did not show a bias towards the answer "All of the above."}
\end{tabularx}
\end{table*}

\begin{itemize}
\item \textbf{BERTology}~\cite{vimmrc2}: This model integrates the pre-trained mBERT \cite{vibert}, a multi-step attention mechanism (MAN), and natural language inference (NLI). Initially, the pre-trained model undergoes fine-tuning using a multi-stage multi-task learning technique \cite{mmm}. Following this, a coarse-tuning stage is employed for the NLI task. Finally, a MAN is implemented as a top-level classifier to compute the attention score between the reading passage and the question-option pairs.

\item \textbf{GPT-3 \& GPT-3.5}~\cite{gpt3}: 
GPT-3 is 10 times larger than any previous non-sparse language model. This model demonstrates strong performance across numerous NLP datasets. GPT-3.5 builds upon this foundation, achieving superior results and surpassing previous SOTA models. They were evaluated without any gradient updates or fine-tuning, relying solely on text interactions.

\item \textbf{Llama 2}~\cite{touvron2023llama2openfoundation}: This is a foundational language model series ranging from 7 billion to 70 billion parameters. It is notable for its training on trillions of tokens, exclusively using publicly accessible datasets, thereby eliminating the need for inaccessible data. This demonstrates the potential of training top-tier models with openly available resources, achieving excellent performance across various tasks.

\item \textbf{BloomZ}~\cite{bloomz}: BloomZ utilizes multitask prompted fine-tuning (MTF) to adapt pre-trained multilingual BLOOM and mT5 model families. This method enhances the models' ability to handle tasks across multiple languages, leveraging English prompts for fine-tuning. It improves performance on both English and non-English tasks, achieving SOTA results in scenarios where no task-specific training data is available.

\item \textbf{Llama3$_{\text{Base}}$, Llama3$_{\text{Instructed}}$, Gemma$_{\text{Base}}$, Gemma$_{\text{Instructed}}$}: our finetuned models on the
Llama 3 and Gemma base and instruction variants.

\end{itemize}


\section{Results and Discussions}

\subsection{Comparison with baselines}

Table \ref{tab:baseline} compares the performance of our fine-tuned models with the baseline models on the ViMMRC dataset, with performance measured by accuracy as per \cite{baseline}. 
While our fine-tuned models significantly outperform BERT-based approaches, we acknowledge the difference in model sizes. Yet, the dramatic improvements we achieved through fine-tuning suggest that our superior results stem from more than just larger parameter counts – our models substantially outperform BERT even when it uses advanced techniques like NLI fine-tuning and multi-step attention.


\subsection{Advantages with small-size models}

It is also worth noting that our fine-tuned models, with relatively small parameter counts, can outperform much larger models such as GPT-3 and GPT-3.5. This demonstrates the efficiency and effectiveness of our fine-tuning process, which allows smaller models to achieve higher accuracy than these larger, more resource-intensive models. It also suggests significant potential for more efficient deployment of their applications in resource-constrained environments.

\subsection{Improvement from fine-tuned models}

Table \ref{tab:finetune} highlights the comparison between the models before and after fine-tuning. We do not evaluate the base models before fine-tuning because they 
do not consistently answer our questions in a zero-shot setting. The accuracy of the base LLMs after fine-tuning is substantially higher than before fine-tuning. This highlights the positive impact of adapting the models specifically to the ViMMRC dataset, leading to improved comprehension and question-answering capabilities in Vietnamese.

Interestingly, the fine-tuned base LLMs achieve better accuracy compared to their instruction-tuned counterparts. The base model Gemma performs significantly better than the instructed version. We speculate that this is because the instructed Gemma is trained exclusively on English data, lacking the multi-language data necessary for optimal performance on the ViMMRC dataset. Meanwhile, the base Llama 3 model also surpasses its instruction-tuned version, though the difference is marginal. This could be because the instructed Llama is primarily optimized for chat abilities and toxicity mitigation, which are less relevant for the machine reading comprehension tasks.

\begin{table*}[t!]
\centering
\caption{\label{tab:error} Error analyses}

\begin{tabularx}{\textwidth}{l|X|X}
\hline
& \textbf{Sample} & \textbf{English translation} \\
\hline
\textbf{Passage 1} &
Bố đi câu về, không một lần nào là chúng tôi không có quà. Mở thúng câu ra là cả một thế giới dưới nước: cà cuống, niềng niễng đực, niềng niễng cái bò nhộn nhạo. Hoa sen đỏ, nhị sen vàng tỏa hương thơm lừng. Những con cá sộp, cá chuối quẫy tóe nước, mắt thao láo... Bố đi cắt tóc về, cũng không lần nào chúng tôi không có quà. Mở hòm dụng cụ ra là cả một thế giới mặt đất: con xập xành, con muỗm to xù, mốc thếch, ngó ngoáy. Hấp dẫn nhất là những con dế lạo xạo trong các vỏ bao diêm: toàn dế đực, cánh xoăn, gáy vang nhà và chọi nhau phải biết. Quà của bố làm anh em tôi giàu quá! &

When Dad went fishing, he never came back without gifts for us. Opening the fishing basket was like revealing a whole underwater world: giant water bugs, male and female damselflies wriggling around. Red lotus flowers and yellow lotus stamens spread a fragrant aroma. Big fish and snakehead fish splashed water, their eyes bulging. When Dad went for a haircut, he never came back without gifts for us. Opening the toolbox was like revealing a whole terrestrial world: giant crickets and clumsy mole crickets wriggling around. The most fascinating were the chirping crickets in the matchboxes: all male crickets with curly wings, chirping loudly and fiercely fighting each other. Dad's gifts made my siblings and me feel so rich!\\
\hline
\textbf{Question} & Những món quà mà bố mang về cho các con nói lên điều gì? 
& 
What do the gifts that father brings home for the children say? 
\\
\hline
\textbf{Answers} & A. Bố rất yêu thương hai anh em. \newline B. Bố rất giỏi đi câu và cắt tóc. \newline C. Bố rất thích đồng quê và sông nước. \newline D. Bố rất thích tặng quà.
& 
A. He loves his children. \newline B. He's good at fishing and hairdressing. \newline C. He loves countryside. \newline D. He likes gifting.
\\
\hline
\textbf{Predictions} & \multicolumn{2}{l}{
\textbf{Correct Answer}: A \quad \quad
\textbf{Llama 3}: C \quad \quad
\textbf{Gemma}: C
} \\
\hline
\multicolumn{3}{c}{}\\
\hline
\textbf{Passage 2} &
Bút chì xanh đỏ\newline
Em gọt hai đầu\newline
Em thử hai màu\newline
Xanh tươi, đỏ thắm. &

Blue and red pencil\newline
I sharpen both ends\newline
I try both colors\newline
Bright green, vivid red.\\
\hline
\textbf{Question} & Bài thơ đã sử dụng thể thơ nào để viết?
& 
What poetic form was used to write the poem?
\\
\hline
\textbf{Answers} & A. Lục bát. \quad B. Tự do. \quad C. Năm chữ. \quad D. Bốn chữ.
& 
A. six-eight. \quad B. free verse. \quad C. five syllables. \quad D. four syllables.
\\
\hline
\textbf{Predictions} & \multicolumn{2}{l}{
\textbf{Correct Answer}: D \quad \quad
\textbf{Llama 3}: A \quad \quad
\textbf{Gemma}: A
} \\
\hline
\multicolumn{3}{c}{}\\
\hline
\textbf{Passage 3} &
Trong tù không rượu cũng không hoa,\newline Cảnh đẹp đêm nay, khó hững hờ.\newline Người ngắm trăng soi ngoài cửa sổ,\newline Trăng nhòm khe cửa ngắm nhà thơ. &

In prison without wine or flowers,\newline
Tonight's beauty, hard to disregard.\newline
Gazing at the moon through the window bars,\newline
The moon peers through, watching the poet's heart\\
\hline
\textbf{Question} & Bài thơ này cho thấy Bác Hồ là người như thế nào?
& 
What does this poem show about Uncle Ho’s character?
\\
\hline
\textbf{Answers} & A. Bác Hồ là người rất yêu nước. \newline B. Bác Hồ rất yêu thiên nhiên. \newline C. Bác Hồ là người rất giản dị. \newline D. Bác Hồ rất dũng cảm.
& 
A. Uncle Ho greatly loved his country. \newline B. Uncle Ho loved nature very much. \newline C. Uncle Ho was a very simple person. \newline D. Uncle Ho was very brave. \\
\hline
\textbf{Predictions} & \multicolumn{2}{l}{
\textbf{Correct Answer}: B \quad \quad
\textbf{Llama 3}: C \quad \quad
\textbf{Gemma}: A
} \\
\hline
\multicolumn{3}{c}{}\\
\hline
\textbf{Passage 4} &
Chiều hôm ấy có một em gái nhỏ đứng áp trán vào tủ kính cửa hàng của Pi-e, nhìn từng đồ vật như muốn kiếm thứ gì. Bỗng em ngửng đầu lên: - Cháu có thể xem chuỗi ngọc lam này không ạ? Pi-e lấy chuỗi ngọc, đưa cho cô bé. Cô bé thốt lên: - Đẹp quá! Xin chú gói lại cho cháu! Pi-e ngạc nhiên: - Ai sai cháu đi mua? - Cháu mua tặng chị cháu nhân lễ Nô-en. Chị đã nuôi cháu từ khi mẹ cháu mất. - Cháu có bao nhiêu tiền? Cô bé mở khăn tay ra, đổ lên bàn một nắm xu: - Cháu đã đập con lợn đất đấy! Pi-e trầm ngâm nhìn cô bé. Rồi vừa lúi húi gỡ mảnh giấy ghi giá tiền, anh vừa hỏi: - Cháu tên gì? - Cháu là Gioan. Anh đưa Gioan chuỗi ngọc gói trong bao lụa đỏ: - Đừng đánh rơi nhé! Cô bé mỉm cười rạng rỡ, chạy vụt đi... &

One afternoon, a little girl stood at Pierre's shop window, examining items as if searching for something. She asked to see a blue necklace. Pierre showed it to her. Delighted, she requested it be wrapped. When asked who sent her, she explained it was a Christmas gift for her sister who raised her after their mother's death. She emptied her savings onto the counter. Pierre, moved, removed the price tag and asked her name. "I am Jeanne." she replied. He then handed her the necklace in a red silk bag, said "Don't lose it on the way home.". She smiled brightly and ran off....\\
\hline
\textbf{Question} & Từ "cháu" trong câu "Cháu là Gioan" đóng vai trò cú pháp gì?
& 
What syntactic role does the word "I" play in the sentence "I am Jeanne"? 
\\
\hline
\textbf{Answers} & A. Danh từ làm chủ ngữ. \newline B. Đại từ làm chủ ngữ. \newline C. Tính từ làm chủ ngữ. \newline D. Động từ làm chủ ngữ.
& 
A. Noun functioning as the subject. \newline B. Pronoun functioning as the subject. \newline C. Adjective functioning as the subject. \newline D. Verb functioning as the subject.
\\
\hline
\textbf{Predictions} & \multicolumn{2}{l}{
\textbf{Correct Answer}: B \quad \quad
\textbf{Llama 3}: A \quad \quad
\textbf{Gemma}: A
} \\
\hline
\multicolumn{3}{c}{}\\
\multicolumn{3}{p{0.95\linewidth}}{\textit{Notes:} Sample 1 tests understanding of implied meanings in dad's gift-giving. Sample 2 assesses recognition of poetic forms. Sample 3 evaluates comprehension of character traits in poetry. Sample 4 examines understanding of grammatical roles in sentences.}
\end{tabularx}
\end{table*}

\subsection{Bias mitigation}
We inferred our fine-tuned Llama 3$_{\text{Base}}$ and Gemma$_{\text{Base}}$ models on the ViMMRC test dataset to identify positive improvements, understand the nature of their performance, and pinpoint areas where they struggle. These models show clear improvements over the BERTology model in terms of bias. Unlike the BERTology model, which tends to choose "All of the above" as the answer \cite{vimmrc2}, our models do not exhibit this bias, as demonstrated in Table \ref{tab:bias}.

\subsection{Error analysis}

We further investigate these models via error analyses, as shown in Table \ref{tab:error}. The main source of errors is the difficulty in identifying the most suitable answer for questions that require reasoning. Reasoning can be particularly challenging for LLMs because it often involves grasping the underlying message of an entire story or identifying the specific type of a poem based on structural characteristics such as word count. For instance, in Sample 2 of Table V, both models failed to correctly identify the poetic form (four syllables) used in the poem, instead incorrectly choosing the "six-eight" option.

Additionally, they have trouble understanding deep figurative language. 
This highlights the complexity of tasks involving nuanced and context-dependent language. As shown in Sample 3, both models struggled to accurately interpret Uncle Ho's character from the poem, with Llama 3 choosing "a very simple person" and Gemma selecting "greatly loved his country" instead of the correct answer that he "loved nature very much".

Our models also sometimes struggle with correctly answering parts of speech questions, indicating a need for better understanding of grammatical nuances. This suggests areas where further fine-tuning and enhancement could improve model performance. For example, in Sample 4, both Llama 3 and Gemma incorrectly identified the word "Cháu" as a noun functioning as the subject, when it actually functions as a pronoun in the given sentence "Cháu là Gioan".

\subsection{Limitation \& Future work}
Though we obtained promising results, we still need to overcome several limitations. 
While quantization techniques alleviate computational resource constraints and enabling local deployment, this may introduce subtle performance trade-offs. Another 
limitation lies in the scope of our evaluation dataset. The experiments were primarily conducted on the original ViMMRC dataset, excluding newer iterations such as ViMMRC 2.0 \cite{vimmrc2} which extend the original dataset to 12th grade. 
We propose the following directions for future research:
\begin{itemize}
\item Implement Chain-of-Thought prompting techniques \cite{CoT} to enhance the models' reasoning capabilities, potentially improving performance on complex MRC tasks.
\item Expand the evaluation on other datasets such as ViMMRC 2.0 \cite{vimmrc2}, offering more diverse and challenging examples for a comprehensive assessment.
\item Develop practical applications leveraging the fine-tuned models, including local question-answering systems, educational tools for automated question generation.
\end{itemize}

\section{Conclusion}
In this study, we fine-tuned two state-of-the-art LLMs, Llama 3 and Gemma, using the QLoRA method on the ViMMRC dataset, which led to significant performance improvements. Our models not only outperformed traditional BERT-based approaches but also surpassed larger models like GPT-3 and GPT-3.5, despite having fewer parameters. These findings highlight the efficiency and potential of QLoRA-based fine-tuning, making it highly suitable for deploying high-performance NLP models in environments with limited computational resources. Through our in-depth analyses, we also demonstrated the adaptability of LLMs to Vietnamese MRC tasks, providing valuable insights for future advancements in low-resource language processing. Moving forward, future work will focus on enhancing model reasoning through Chain-of-Thought prompting, evaluating performance on newer datasets such as ViMMRC 2.0, and developing practical real-world applications. Overall, this study underscores the potential of scaling LLM-based solutions for underrepresented languages, which can drive further innovation in natural language processing.

\section*{Acknowledgment}
We would like to thank the anonymous reviewers for their valuable feedback. This work was supported in part by computing resources provided by \href{https://www.int2.vn/}{Intelligent Integration Co., Ltd. (INT2)}.

\bibliography{kse24}

\begin{thebibliography}{10}
\providecommand{\url}[1]{#1}
\csname url@samestyle\endcsname
\providecommand{\newblock}{\relax}
\providecommand{\bibinfo}[2]{#2}
\providecommand{\BIBentrySTDinterwordspacing}{\spaceskip=0pt\relax}
\providecommand{\BIBentryALTinterwordstretchfactor}{4}
\providecommand{\BIBentryALTinterwordspacing}{\spaceskip=\fontdimen2\font plus
\BIBentryALTinterwordstretchfactor\fontdimen3\font minus \fontdimen4\font\relax}
\providecommand{\BIBforeignlanguage}[2]{{%
\expandafter\ifx\csname l@#1\endcsname\relax
\typeout{** WARNING: IEEEtran.bst: No hyphenation pattern has been}%
\typeout{** loaded for the language `#1'. Using the pattern for}%
\typeout{** the default language instead.}%
\else
\language=\csname l@#1\endcsname
\fi
#2}}
\providecommand{\BIBdecl}{\relax}
\BIBdecl

\bibitem{dzendzik-etal-2021-english}
\BIBentryALTinterwordspacing
D.~Dzendzik, J.~Foster, and C.~Vogel, ``{E}nglish machine reading comprehension datasets: A survey,'' in \emph{Proceedings of the 2021 Conference on Empirical Methods in Natural Language Processing}, M.-F. Moens, X.~Huang, L.~Specia, and S.~W.-t. Yih, Eds.\hskip 1em plus 0.5em minus 0.4em\relax Online and Punta Cana, Dominican Republic: Association for Computational Linguistics, Nov. 2021, pp. 8784--8804. [Online]. Available: \url{https://aclanthology.org/2021.emnlp-main.693}
\BIBentrySTDinterwordspacing

\bibitem{9664302}
P.~Zhu, Z.~Zhang, H.~Zhao, and X.~Li, ``Duma: Reading comprehension with transposition thinking,'' \emph{IEEE/ACM Transactions on Audio, Speech, and Language Processing}, vol.~30, pp. 269--279, 2022.

\bibitem{Zhao2023ASO}
\BIBentryALTinterwordspacing
W.~X. Zhao, K.~Zhou, J.~Li, T.~Tang, X.~Wang, Y.~Hou, Y.~Min, B.~Zhang, J.~Zhang, Z.~Dong, Y.~Du, C.~Yang, Y.~Chen, Z.~Chen, J.~Jiang, R.~Ren, Y.~Li, X.~Tang, Z.~Liu, P.~Liu, J.~Nie, and J.~rong Wen, ``A survey of large language models,'' \emph{ArXiv}, vol. abs/2303.18223, 2023. [Online]. Available: \url{https://api.semanticscholar.org/CorpusID:257900969}
\BIBentrySTDinterwordspacing

\bibitem{Zhou2023ACS}
\BIBentryALTinterwordspacing
C.~Zhou, Q.~Li, C.~Li, J.~Yu, Y.~Liu, G.~Wang, K.~Zhang, C.~Ji, Q.~Yan, L.~He, H.~Peng, J.~Li, J.~Wu, Z.~Liu, P.~Xie, C.~Xiong, J.~Pei, P.~S. Yu, L.~S. M.~S. University, B.~University, L.~University, M.~University, N.~T. University, U.~of~California~at San~Diego, D.~University, U.~of~Chicago, and S.~Research, ``A comprehensive survey on pretrained foundation models: A history from bert to chatgpt,'' \emph{ArXiv}, vol. abs/2302.09419, 2023. [Online]. Available: \url{https://api.semanticscholar.org/CorpusID:257039063}
\BIBentrySTDinterwordspacing

\bibitem{Bommasani2021FoundationModels}
\BIBentryALTinterwordspacing
R.~Bommasani, D.~A. Hudson, E.~Adeli, R.~Altman, S.~Arora, S.~von Arx, M.~S. Bernstein, J.~Bohg, A.~Bosselut, E.~Brunskill, E.~Brynjolfsson, S.~Buch, D.~Card, R.~Castellon, N.~S. Chatterji, A.~S. Chen, K.~A. Creel, J.~Davis, D.~Demszky, C.~Donahue, M.~Doumbouya, E.~Durmus, S.~Ermon, J.~Etchemendy, K.~Ethayarajh, L.~Fei-Fei, C.~Finn, T.~Gale, L.~E. Gillespie, K.~Goel, N.~D. Goodman, S.~Grossman, N.~Guha, T.~Hashimoto, P.~Henderson, J.~Hewitt, D.~E. Ho, J.~Hong, K.~Hsu, J.~Huang, T.~F. Icard, S.~Jain, D.~Jurafsky, P.~Kalluri, S.~Karamcheti, G.~Keeling, F.~Khani, O.~Khattab, P.~W. Koh, M.~S. Krass, R.~Krishna, R.~Kuditipudi, A.~Kumar, F.~Ladhak, M.~Lee, T.~Lee, J.~Leskovec, I.~Levent, X.~L. Li, X.~Li, T.~Ma, A.~Malik, C.~D. Manning, S.~P. Mirchandani, E.~Mitchell, Z.~Munyikwa, S.~Nair, A.~Narayan, D.~Narayanan, B.~Newman, A.~Nie, J.~C. Niebles, H.~Nilforoshan, J.~F. Nyarko, G.~Ogut, L.~Orr, I.~Papadimitriou, J.~S. Park, C.~Piech, E.~Portelance, C.~Potts, A.~Raghunathan, R.~Reich, H.~Ren, F.~Rong, Y.~H. Roohani,
  C.~Ruiz, J.~Ryan, C.~R'e, D.~Sadigh, S.~Sagawa, K.~Santhanam, A.~Shih, K.~P. Srinivasan, A.~Tamkin, R.~Taori, A.~W. Thomas, F.~Tram{\`e}r, R.~E. Wang, W.~Wang, B.~Wu, J.~Wu, Y.~Wu, S.~M. Xie, M.~Yasunaga, J.~You, M.~A. Zaharia, M.~Zhang, T.~Zhang, X.~Zhang, Y.~Zhang, L.~Zheng, K.~Zhou, and P.~Liang, ``On the opportunities and risks of foundation models,'' \emph{ArXiv}, 2021. [Online]. Available: \url{https://crfm.stanford.edu/assets/report.pdf}
\BIBentrySTDinterwordspacing

\bibitem{Gholami2023DoGL}
\BIBentryALTinterwordspacing
S.~Gholami and M.~Omar, ``Do generative large language models need billions of parameters?'' \emph{ArXiv}, vol. abs/2309.06589, 2023. [Online]. Available: \url{https://api.semanticscholar.org/CorpusID:261705636}
\BIBentrySTDinterwordspacing

\bibitem{vimmrc}
\BIBentryALTinterwordspacing
K.~V. Nguyen, K.~V. Tran, S.~T. Luu, A.~G.-T. Nguyen, and N.~L.-T. Nguyen, ``A pilot study on multiple choice machine reading comprehension for vietnamese texts,'' \emph{ArXiv}, vol. abs/2001.05687, 2020. [Online]. Available: \url{https://api.semanticscholar.org/CorpusID:210700877}
\BIBentrySTDinterwordspacing

\bibitem{llama3modelcard}
\BIBentryALTinterwordspacing
AI@Meta, ``Llama 3 model card,'' 2024. [Online]. Available: \url{https://github.com/meta-llama/llama3/blob/main/MODEL_CARD.md}
\BIBentrySTDinterwordspacing

\bibitem{team2024gemma}
G.~Team, T.~Mesnard, C.~Hardin, R.~Dadashi, S.~Bhupatiraju, S.~Pathak, L.~Sifre, M.~Rivi{\`e}re, M.~S. Kale, J.~Love \emph{et~al.}, ``Gemma: Open models based on gemini research and technology,'' \emph{arXiv preprint arXiv:2403.08295}, 2024.

\bibitem{touvron2023llama2openfoundation}
\BIBentryALTinterwordspacing
H.~Touvron, L.~Martin, K.~Stone, P.~Albert, A.~Almahairi, Y.~Babaei, N.~Bashlykov, S.~Batra, P.~Bhargava, S.~Bhosale, D.~Bikel, L.~Blecher, C.~C. Ferrer, M.~Chen, G.~Cucurull, D.~Esiobu, J.~Fernandes, J.~Fu, W.~Fu, B.~Fuller, C.~Gao, V.~Goswami, N.~Goyal, A.~Hartshorn, S.~Hosseini, R.~Hou, H.~Inan, M.~Kardas, V.~Kerkez, M.~Khabsa, I.~Kloumann, A.~Korenev, P.~S. Koura, M.-A. Lachaux, T.~Lavril, J.~Lee, D.~Liskovich, Y.~Lu, Y.~Mao, X.~Martinet, T.~Mihaylov, P.~Mishra, I.~Molybog, Y.~Nie, A.~Poulton, J.~Reizenstein, R.~Rungta, K.~Saladi, A.~Schelten, R.~Silva, E.~M. Smith, R.~Subramanian, X.~E. Tan, B.~Tang, R.~Taylor, A.~Williams, J.~X. Kuan, P.~Xu, Z.~Yan, I.~Zarov, Y.~Zhang, A.~Fan, M.~Kambadur, S.~Narang, A.~Rodriguez, R.~Stojnic, S.~Edunov, and T.~Scialom, ``Llama 2: Open foundation and fine-tuned chat models,'' 2023. [Online]. Available: \url{https://arxiv.org/abs/2307.09288}
\BIBentrySTDinterwordspacing

\bibitem{qlora}
\BIBentryALTinterwordspacing
T.~Dettmers, A.~Pagnoni, A.~Holtzman, and L.~Zettlemoyer, ``Qlora: Efficient finetuning of quantized llms,'' 2023. [Online]. Available: \url{https://arxiv.org/abs/2305.14314}
\BIBentrySTDinterwordspacing

\bibitem{Unsloth}
\BIBentryALTinterwordspacing
D.~Han and M.~Han, ``{Unsloth AI},'' 5 2024. [Online]. Available: \url{https://github.com/unslothai/unsloth}
\BIBentrySTDinterwordspacing

\bibitem{vimmrc2}
\BIBentryALTinterwordspacing
S.~T. Luu, K.~T. Hoang, T.~Q. Pham, K.~V. Nguyen, and N.~L.-T. Nguyen, ``A multiple choices reading comprehension corpus for vietnamese language education,'' 2023. [Online]. Available: \url{https://arxiv.org/abs/2303.18162}
\BIBentrySTDinterwordspacing

\bibitem{vibert}
\BIBentryALTinterwordspacing
V.~B. The, O.~T. Thi, and P.~Le-Hong, ``Improving sequence tagging for vietnamese text using transformer-based neural models,'' \emph{Pacific Asia Conference on Language, Information and Computation}, 2020. [Online]. Available: \url{https://arxiv.org/abs/2006.15994}
\BIBentrySTDinterwordspacing

\bibitem{mmm}
\BIBentryALTinterwordspacing
D.~Jin, S.~Gao, J.-Y. Kao, T.~Chung, and D.~Z. Hakkani-T{\"u}r, ``Mmm: Multi-stage multi-task learning for multi-choice reading comprehension,'' in \emph{AAAI Conference on Artificial Intelligence}, 2019. [Online]. Available: \url{https://api.semanticscholar.org/CorpusID:203610341}
\BIBentrySTDinterwordspacing

\bibitem{gpt3}
\BIBentryALTinterwordspacing
T.~B. Brown, B.~Mann, N.~Ryder, M.~Subbiah, J.~Kaplan, P.~Dhariwal, A.~Neelakantan, P.~Shyam, G.~Sastry, A.~Askell, S.~Agarwal, A.~Herbert-Voss, G.~Krueger, T.~Henighan, R.~Child, A.~Ramesh, D.~M. Ziegler, J.~Wu, C.~Winter, C.~Hesse, M.~Chen, E.~Sigler, M.~Litwin, S.~Gray, B.~Chess, J.~Clark, C.~Berner, S.~McCandlish, A.~Radford, I.~Sutskever, and D.~Amodei, ``Language models are few-shot learners,'' 2020. [Online]. Available: \url{https://arxiv.org/abs/2005.14165}
\BIBentrySTDinterwordspacing

\bibitem{bloomz}
\BIBentryALTinterwordspacing
N.~Muennighoff, T.~Wang, L.~Sutawika, A.~Roberts, S.~Biderman, T.~L. Scao, M.~S. Bari, S.~Shen, Z.-X. Yong, H.~Schoelkopf, X.~Tang, D.~Radev, A.~F. Aji, K.~Almubarak, S.~Albanie, Z.~Alyafeai, A.~Webson, E.~Raff, and C.~Raffel, ``Crosslingual generalization through multitask finetuning,'' 2023. [Online]. Available: \url{https://arxiv.org/abs/2211.01786}
\BIBentrySTDinterwordspacing

\bibitem{baseline}
\BIBentryALTinterwordspacing
D.-V. Nguyen and Q.-N. Nguyen, ``Evaluating the symbol binding ability of large language models for multiple-choice questions in vietnamese general education,'' 2023. [Online]. Available: \url{https://arxiv.org/abs/2310.12059}
\BIBentrySTDinterwordspacing

\bibitem{CoT}
\BIBentryALTinterwordspacing
J.~Wei, X.~Wang, D.~Schuurmans, M.~Bosma, B.~Ichter, F.~Xia, E.~Chi, Q.~Le, and D.~Zhou, ``Chain-of-thought prompting elicits reasoning in large language models,'' 2022. [Online]. Available: \url{https://arxiv.org/pdf/2201.11903}
\BIBentrySTDinterwordspacing

\end{thebibliography}
\bibliographystyle{IEEEtran}

\end{document}